\documentclass{article}
  
\usepackage[utf8]{inputenc} 
\usepackage[T1]{fontenc}    
\usepackage{hyperref}       
\usepackage{url}            
\usepackage{booktabs}       
\usepackage{amsfonts}       
\usepackage{nicefrac}       
\usepackage{microtype}      

\title{A New Compensatory Genetic Algorithm-Based Method for Effective Compressed Multi-function Convolutional Neural Network Model Selection with Multi-Objective Optimization}

\author{%
  Luna M. Zhang
}

\begin{document}

\maketitle

\begin{abstract}
  In recent years, there have been many popular Convolutional Neural Networks (CNNs), such as Google's Inception-V4, that have performed very well for various image classification problems. These commonly used CNN models usually use the same activation function, such as RELU, for all neurons in the convolutional layers; they are ``Single-function CNNs." However, SCNNs may not always be optimal for different image classification problems with different image data in terms of image classification accuracy for real-world applications. Thus, a ``Multi-function CNN" (MCNN), which uses different activation functions for different neurons, has been shown to outperform a SCNN. Also, CNNs typically have very large architectures that use a lot of memory and need a lot of data in order to be trained well. As a result, they tend to have very high training and prediction times too. A smaller CNN model takes less training time, less prediction time, less power, and less memory than a larger one. An important research problem is how to automatically and efficiently find the best CNN with both high classification performance and compact architecture with high training and prediction speeds, small power usage, and small memory size for any image classification problem. It is very useful to intelligently find an effective, fast, energy-efficient, and memory-efficient ``Compressed Multi-function CNN" (CMCNN) from a large number of candidate MCNNs. A new compensatory algorithm using a new genetic algorithm (GA) is created to find the best CMCNN with an ideal compensation between performance and architecture size. The optimal CMCNN has the best performance and the smallest architecture size. Simulations using the CIFAR10 dataset showed that the new compensatory algorithm could find CMCNNs that could outperform non-compressed MCNNs in terms of classification performance (F1-score), speed, power usage, and memory usage. Other effective, fast, power-efficient, and memory-efficient CMCNNs based on popular CNN architectures will be developed and optimized for image classification problems in important real-world applications, such as brain informatics and biomedical imaging.
\end{abstract}

\section{Introduction}

Recently, deep learning techniques have been effectively used in various applications in computer vision, healthcare, etc. Convolutional Neural Networks (CNNs) are very effective techniques for image classification for various important real-world applications [1-6]. Some examples are GoogLeNet, ResNets, DenseNets, Dual Path Networks, and Inception-v4 networks. Traditional CNNs tend to use the same activation function (typically RELU) for all convolutional layers. For example, both ResNets [3] and the very deep Inception-v4 network [6] use the RELU for their activation functions. However, CNNs with many layers and neurons and using RELU everywhere may not always be optimal in terms of image classification performance (i.e. accuracy, F1-score), training time, power efficiency, and memory efficiency. Compared to smaller networks, larger networks would need more memory, longer training times (less power-efficient), longer prediction times, and more data to perform and generalize well. Therefore, it would be advantageous to try to make CNNs smaller in size.  Thus, an important and interesting research problem is how to automatically find the best hyperparameters and small CNN architectures with the best classification performance, fast training time, small power usage, and small memory usage. 

Neural architecture search (NAS) is an important subfield of automated machine learning (AutoML). Much work has already been done in NAS and AutoML. In particular, evolutionary approaches, such as Genetic Algorithms (GAs), have been used for NAS. For example, GA was used to optimize the number of layers and the number of neurons of each layer of a CNN that used one activation function [7, 8]. Also, a multi-objective GA was used for NAS by proposing a new algorithm called NSGA-Net [9]. The results for both examples showed that using GA could be useful for optimizing a CNN's architecture and a CNN's performance at the same time. 

Deep learning models like CNNs, such as Inception-V4, tend to be large in size with a lot of parameters. As a result, they use a lot of memory and need a lot of data in order to be trained well. Also, they tend to have very high training and prediction times too. Some work has been done in pruning CNNs to be smaller [11]. There are many benefits to remove parameters, neurons, or layers that are redundant or do not contribute much to the final output. This is especially important for running deep neural networks on mobile devices that have limited memory and computing power.

A``Multi-function CNN" (MCNN) uses different activation functions for different convolutional layers. MCNNs have been shown to perform better than CNNs using RELU. A ``Compressed Multi-function Convolutional Neural Network" (CMCNN) uses a reduced number of convolutional layers (and activation functions) than the original CNN used. There are many choices for the activation functions and the number of convolutional layers, so we design a new multi-objective GA to automatically optimize CMCNNs and select the best CMCNN that takes into account both the image classification performance and architecture size. The goal is to find the best CMCNN with high performance, small architecture size, and fast prediction time. Such an effective, fast, power-efficient, and memory-efficient CMCNN is useful for practical applications, such as AI chips, and mobile biomedical imaging.

\section{Multi-function CNNs and Compressed Multi-function CNNs}

A CNN consists of convolutional, activation, pooling, and fully connected layers. An activation layer is a layer of neurons where each neuron uses an activation function. A typical CNN uses a RELU layer right after each convolutional layer and before a pooling layer to generate new non-linear feature maps from the initial feature maps. Different activation functions can be used to build a ``Multi-function CNN" (MCNN). 

For a popular CNN, such as Inception-V4 by Google, it would be useful to see if making its architecture smaller (i.e. removing layers) and making it ``multi-function" can lead to better performance, faster speeds in training and prediction, lower energy consumption, and lower memory usage for a particular image classification problem. 

To show the feasibility of finding CMCNNs that can perform better than an original CNN, Google’s Inception-V4 architecture was used as the base framework. For testing purposes, the numbers of Inception-A blocks, Inception-B blocks, and
Inception-C blocks were reduced. 
Let "$CMI_{i}$" and "$CI_{i}$" mean that a CMI and a
compressed Inception-V4 with RELU have $i$ Inception-A
block(s), $i$ + 1 Inception-B block(s), and $i$ Inception-C
block(s) for $i$ = 1, 2, and 3. For $CMI_{1}$, the number of activation
functions is 58, $CMI_{2}$
has 85 activation functions, and $CMI_{3}$ has 112. In Tables 1 and 2, "MI" and “I” mean that a MI
and the original Inception-V4 with RELU have 4 Inception-
A, 7 Inception-B, and 3 Inception-C blocks. Stratified 3-
fold cross validation was used to evaluate and compare the three CMI models, the MI model, the three compressed
Inception-V4 models with RELU, and the original Inception-V4 using multi-class classification metrics (i.e. training
F1-score ($F1_{train}$), validation F1-scores ($F1_{valid}$), training
times ($T_{train}$) in seconds, and classification testing times
($T_{test}$) in seconds. An activation function set {RELU, SIG,
TANH, ELU} was used to build all of the MCNNs. Each activation function is randomly chosen from
this set for each convolutional block.
Table 1 shows the number of convolutional blocks (CBs), model
sizes, and numbers of activation functions used for $CMI_{1}$,
$CMI_{2}$, $CMI_{3}$, and $MI$.

\begin{table}[h!]
  \caption{Different model architectures}
  \label{sample-table}
  \centering
  \begin{tabular}{lllll}
    \toprule
Model: & $CMI_{1}$ & $CMI_{2}$ & $CMI_{3}$  & $MI$  \\
    \midrule
No. CBs & 58& 85& 112 & 149 \\
Model Size (MB) & 129& 190& 252 & 323\\
No. Functions & 58& 85&112 & 149\\
    \bottomrule
  \end{tabular}
\end{table}
 
Some initial simulations have been done to show that removing some layers from Inception-V4 and adding in the "multi-function" feature can improve performance. A dataset of 436 brain MRI images (cross-sectional collection
of 416 subjects aged 18 to 96 and with extra data for
20 subjects) is used for
performance analysis [13]. All brain MRI images for a 4-class classification problem to determine
the Alzheimer’s Disease stage (non-demented, very mild dementia,
mild dementia, or moderate dementia) of a person was used
[13][14]. Table 2 shows that the original Inception-V4 ($I$) has the lowest training F1-score and testing F1-score. The 3 compressed models performed better and faster than the original Inception-V4. Thus, it is feasible to find a CMCNN that may outperform the original CNN. Now the goal is to find an ideal CMCNN among a huge number of candidate CMCNNs to achieve multiple objectives at the same time, such as best performance (i.e. classification accuracy), fastest inference speed, lowest energy usage, and lowest memory usage.

\begin{table}[h!]
  \caption{Model comparisons (436 brain images, stratified 3-fold cross validation, 120 epochs)}
  \label{sample-table}
  \centering
  \begin{tabular}{llllll}
    \toprule
Model: & $CMI_{1}$ & $CMI_{2}$ & $CMI_{3}$  & $MI$   & $I$   \\
    \midrule
$F1_{train}$ & 0.78& 0.85& \textbf{0.86} & 0.84& 0.70 \\
$F1_{test}$ & 0.76& 0.82& \textbf{0.86} & 0.83& 0.68\\
$T_{train} (s)$ & 2229& 3081& 3978 & 5082& 4815 \\
$T_{test} (s)$ & 1.24& 1.54& 1.88 & 2.41& 2.31 \\
    \bottomrule
  \end{tabular}
\end{table}

\section{New compensatory CMCNN model selection algorithms using GA for multi-objective optimization}

\subsection{CNN model selection with multi-objective optimization}

We consider a four-objective optimization problem: maximizing performance and speed, and minimizing energy usage and memory usage by optimizing CMCNN’s hyperparameters, such as the numbers of layers, numbers of neurons in layers, and activation functions of neurons. 

The speed of a computer is defined as the number of instructions per second. A computer speed is denoted as $S$. The power (in Joules) is $E=CTS^\alpha$, where $C$ and $\alpha$ are constants [12]. Then, the energy $E$ is directly proportional to the execution time $T$. If the number of convolutional layers of a CMCNN is reduced, then the execution time and energy consumption are also reduced. In addition, a CNN model size (i.e., memory usage) is reduced too. 
Thus, the four-objective optimization problem becomes a two-objective optimization problem: maximzing performance and minimizing the number of convolutional layers of a CMCNN.   

A simple optimization function is defined as $\alpha = wF + (1 - w)(1 - S)$ where $w$ for $0\leq w\leq 1$ is a weight, and $S$ is a metric related to a property of a CMCNN architecture, such as a ratio of the model sizes between a CMCNN and a popular CNN. A user can choose a value for $w$.

\subsection{New compensatory CMCNN model selection algorithms using GA for multi-objective optimization}

A MCNN has $n$ convolutional layers; each convolutional layer is followed by $f\in\{f_1, f_2, ..., f_m\}$ where $f_i$ is an activation function for $i=1, 2, ..., m$. There are $m^n-m$ different MCNNs and $m$ different traditional CNNs that use the same activation function for all neurons. Since there are too many different MCNNs, it is inefficient and unnecessary to test each one. Thus, we develop a new GA. A MCNN's activation functions can be represented by a string $[g_1g_2g_3......g_n]$ where $g_i\in\{f_1, f_2, ..., f_m\}$ for $i=1, 2, ..., n$. The new GA method has two operations: a newly defined mutation and a traditional crossover. For a given string $[a_1a_2a_3......a_n]$ where $a_i\in\{f_1, f_2, ..., f_m\}$ for $i=1, 2, ..., n$, a new mutation operation with a mutation point $j$ for $j\in\{1, 2, 3, ..., n\}$ results in $[a_1a_2...a_{j-1}c_ja_{j+1}...a_n]$ where $c_j\in\{f_1, f_2, ..., f_m\}$ and $c_j \neq a_j$. For two parent strings $[a_1a_2a_3......a_n]$ where $a_i\in\{f_1, f_2, ..., f_m\}$ and $[b_1b_2b_3......b_n]$ where $b_i\in\{f_1, f_2, ..., f_m\}$ for $i=1, 2, ..., n$, a crossover operation with a crossover point $k$ for $k\in\{2, 3, ..., n-1\}$ results in two new child strings $[a_1a_2...a_kb_{k+1}b_{k+2}...b_n]$ and $[b_1b_2...b_ka_{k+1}a_{k+2}...a_n]$.  The new GA method is shown in Algorithm 1. A new CMCNN model selection method using a new GA is shown in Algorithm 2. The new GA using the new mutation operator for optimizing MCNN models is shown in Algorithm 2. $F$ is the fitness function for GA, such as F1-score.

\textbf {Algorithm 1: A New GA for Optimizing MCNN Models}  \newline
 \textbf {Input}: initial population of $N$ MCNN models where $N$ is an even number, training data  \newline
 \textbf {Output}: the best MCNN model  \newline
      \textbf {Step 1.} Create a population of an even number of MCNN models with randomly generated activation functions where each function has equal probability of being chosen. \newline
      \textbf {Step 2.} Train all MCNN models in the population. \newline
      \textbf {Step 3.} Compute $F$ for each trained MCNN model in the population. Keep the top MCNN with the highest $F$. \newline
\textbf {Step 4.} Select pairs of MCNNs' function strings. \newline
      \textbf {Step 5.} Perform crossover on each pair of MCNNs' function strings to generate new MCNNs' function strings. \newline
     \textbf {Step 6.} Perform the newly defined mutation operations on the new MCNNs' activation function strings based on the mutation probability. \newline
     \textbf {Step 7.} Create a new population that includes the newly created MCNNs' activation function strings and the parents. \newline
      \textbf {Step 8.} Train all MCNN models  in the new population. \newline
      \textbf {Step 9.} Compute $F$ for each trained MCNN model in the new population. \newline
\textbf {Step 10.} Repeat steps 4 to 9 for a generation until the maximum number of generations is reached.

\textbf {Algorithm 2: the Compensatory CMCNN Model Selection Using Algorithm 1} \newline
 \textbf {Input}: initial population of $N$ MCNN models where $N$ is an even number, training data  \newline
 \textbf {Output}: the best MCNN model  \newline
      \textbf {Step 1.} For each compressed architecture, create an initial population with $N$ CMCNNs for an even positive integer $N$. \newline
      \textbf {Step 2.} For each compressed architecture, run the new GA for M generations for a positive integer $M$ to find the best CMCNN with the highest $F$. Calculate $\alpha$ for this best CMCNN. \newline
      \textbf {Step 3.} Choose the overall best compensatory CMCNN with the highest $\alpha$ among all the best CMCNNs of different compressed architectures. \newline
  \textbf {Step 4.} Use the overall best compensatory CMCNN for a real-world application.

\section{Simulation results and performance analysis} 

Let ``CLs" mean convolutional layers. Let ``$CM^{n}$" mean that a CMCNN has $n$ CLs with $n$ activation functions. Let ``$CM_{GA}^{n}$" mean that a CMCNN optimized by the new GA has $n$ CLs with $n$ activation functions. Let ``$M^{m}$" mean that a non-compressed MCNN has $m$ CLs with $m$ activation functions for $m > n$. Then we define the model size ratio as $S$ = (number of CLs)/$m$. A fitness function $\alpha = wF + (1 - w)(1 - S)$ for $w = 0.7$ was used for simulations. An activation function set \{RELU, SIG, TANH, ELU\} was used; each function was randomly chosen from this set for each CL. $CM_{GA}^{4}$, $CM_{GA}^{6}$, and $CM_{GA}^{8}$ were compared with $M^{10}$. Samples of the CIFAR10 data were used [10]. The population of the new GA has 4 randomly generated CNNs. The mutation probability is 1. $F$ is the training F1-score. Table 1 shows the numbers of convolutional layers, model sizes (in KB), and average training times (Avg. $T_{train}$) (in seconds) of one CNN model of $CM_{GA}^{4}$, $CM^{4}$, $CM_{GA}^{6}$, $CM^{6}$, $CM_{GA}^{8}$, $CM^{8}$, $M_{GA}^{10}$, and $M^{10}$ for 25000 training data. Avg. $T_{predict} (s)$ is the total prediction time for 7000 testing data. Table 3 shows the properties of some different models. Table 3 shows that CMCNNs use less memory, less training times (i.e., less power usage), less prediction times, and smaller numbers of CLs than $M^{10}$ (a MCNN). $CM_{GA}^{k}$ has longer training time than $CM^{k}$ for $k=4, 6, 8$, and $M_{GA}^{10}$ also has longer training time than $M^{10}$. 

\begin{table}[h!]
  \caption{Properties of different models}
  \label{sample-table}
  \centering
  \begin{tabular}{lllllllll}
    \toprule
    Model: & $CM_{GA}^{4}$ & $CM^{4}$ & $CM_{GA}^{6}$  & $CM^{6}$   & $CM_{GA}^{8}$   & $CM^{8}$ & $M_{GA}^{10}$   & $M^{10}$ \\
    \midrule
      No. of CLs & 4 & 4 & 6 &6&8  &8 &10&10\\
  Model Size (KB)& 454& 454 & 507& 507 & 648& 648&815&815\\
  $S$ & 0.4& 0.4 & 0.6& 0.6 & 0.8& 0.8&1.0&1.0\\
  Avg. $T_{train}(s)$ & 7371&1256 & 9126&1472&9849 &1578 & 10391&1652\\
    Avg. $T_{predict}(s)$ &3.68 &3.68 & 4.04 &4.04& 4.23& 4.23&4.46 &4.46\\
    \bottomrule
  \end{tabular}
\end{table}

\subsection{Simulation results}

Algorithm 2 was implemented. For Tables 4-18, the performance results (training F1-score ($F1_{train}$), testing F1-score ($F1_{test}$), training $\alpha$ ($Fit_{train}$), and testing $\alpha$ ($Fit_{test}$)) of the best models (those with the largest $F1_{train}$) were recorded. The new GA ran many generations of training with new mutation operations and crossover operations. To better evaluate the new algorithm, two CIFAR10 data partitions were made based on the original 50,000 CIFAR10 training data (D[1], D[2], …, D[50000]) and 10,000 CIFAR10 testing data (T[1], T[2], …, T[10000]). The first data partition method generated $N_{train}$ training data (D[1], D[2], …, D[$N_{train}$]) and $N_{test}$ testing data (T[1], T[2], …, T[$N_{test}$]). The second data partition method generated $K_{train}$ training data (D[$50001-K_{train}$], D[$50002-K_{train}$], …, D[50000]) and $K_{test}$ testing data (T[$10001-K_{test}$], T[$10002-K_{test}$], …, T[10000]).

\subsubsection{Simulation results using the first data partition method}

Simulation results using the first data partition method are shown in Tables 4-13.

\begin{table}[h!]
  \caption{Model comparisons ($N_{train} = 20000$, $N_{test} = 5000$, 15 epochs, 5 generations)}
  \label{sample-table}
  \centering
  \begin{tabular}{lllllllll}
    \toprule
Model: & $CM_{GA}^{4}$ & $CM^{4}$ & $CM_{GA}^{6}$  & $CM^{6}$   & $CM_{GA}^{8}$   & $CM^{8}$ & $M_{GA}^{10}$   & $M^{10}$ \\
    \midrule
$F1_{train}$ & \textbf{0.852}& 0.826& 0.829 & 0.810& 0.806 & 0.778&0.770&0.752\\
$F1_{test}$ & 0.737& 0.707& \textbf{0.747} & 0.742& 0.746 & 0.714&0.717&0.714\\
$Fit_{train}$ & \textbf{0.776}& 0.758& 0.700 & 0.687& 0.624 & 0.605&0.539&0.526\\
$Fit_{test}$ & \textbf{0.696}& 0.675& 0.643 & 0.639& 0.582 & 0.560&0.502&0.500\\
    \bottomrule
  \end{tabular}
\end{table}

\begin{table}[h!]
  \caption{Model comparisons ($N_{train} = 25000$, $N_{test} = 6000$, 15 epochs, 5 generations)}
  \label{sample-table}
  \centering
  \begin{tabular}{lllllllll}
    \toprule
Model: & $CM_{GA}^{4}$ & $CM^{4}$ & $CM_{GA}^{6}$  & $CM^{6}$   & $CM_{GA}^{8}$   & $CM^{8}$ & $M_{GA}^{10}$   & $M^{10}$ \\
    \midrule
$F1_{train}$ & \textbf{0.846}& 0.799& 0.829 & 0.813& 0.805 & 0.792&0.732&0.684\\
$F1_{test}$ & 0.734& 0.711& \textbf{0.757} & 0.741& 0.743 & 0.730&0.700&0.654\\
$Fit_{train}$ & \textbf{0.772}& 0.739& 0.700 & 0.689& 0.624 & 0.614&0.512&0.479\\
$Fit_{test}$ & \textbf{0.694}& 0.678& 0.650 & 0.639& 0.580 & 0.571&0.490&0.458\\
    \bottomrule
  \end{tabular}
\end{table}

\begin{table}[h!]
  \caption{Model comparisons ($N_{train} = 12500$, $N_{test} = 5000$, 10 epochs, 10 generations)}
  \label{sample-table}
  \centering
  \begin{tabular}{lllllllll}
    \toprule
Model: & $CM_{GA}^{4}$ & $CM^{4}$ & $CM_{GA}^{6}$  & $CM^{6}$   & $CM_{GA}^{8}$   & $CM^{8}$ & $M_{GA}^{10}$   & $M^{10}$ \\
    \midrule
$F1_{train}$ & \textbf{0.842}& 0.831 &  0.804& 0.775& 0.770 & 0.691&0.689&0.657\\
$F1_{test}$ & 0.697& 0.685 & \textbf{0.712} & 0.690& 0.693 & 0.637&0.640&0.615\\
$Fit_{train}$ & \textbf{0.769}& 0.761& 0.683 & 0.663& 0.599 & 0.544&0.482&0.460\\
$Fit_{test}$ & \textbf{0.668}& 0.660& 0.618 & 0.603& 0.545 & 0.506&0.448&0.431\\
    \bottomrule
  \end{tabular}
\end{table}

\begin{table}[h!]
  \caption{Model comparisons ($N_{train} = 15000$, $N_{test} = 6000$, 20 epochs, 10 generations)}
  \label{sample-table}
  \centering
  \begin{tabular}{lllllllll}
    \toprule
Model: & $CM_{GA}^{4}$ & $CM^{4}$ & $CM_{GA}^{6}$  & $CM^{6}$   & $CM_{GA}^{8}$   & $CM^{8}$ & $M_{GA}^{10}$   & $M^{10}$ \\
    \midrule
$F1_{train}$ & \textbf{0.899}& 0.895& 0.865 & 0.839& 0.801 & 0.801&0.785&0.739\\
$F1_{test}$ & 0.694& 0.695& \textbf{0.740} & 0.722& 0.709 & 0.709&0.710&0.681\\
$Fit_{train}$ & \textbf{0.809}& 0.807& 0.726 & 0.707& 0.621 & 0.621&0.550&0.517\\
$Fit_{test}$ & 0.666& \textbf{0.667}& 0.638 & 0.625& 0.556 & 0.556&0.497&0.477\\
    \bottomrule
  \end{tabular}
\end{table}

\begin{table}[h!]
  \caption{Model comparisons ($N_{train} = 20000$, $N_{test} = 7000$, 20 epochs, 15 generations)}
  \label{sample-table}
  \centering
  \begin{tabular}{lllllllll}
    \toprule
Model: & $CM_{GA}^{4}$ & $CM^{4}$ & $CM_{GA}^{6}$  & $CM^{6}$   & $CM_{GA}^{8}$   & $CM^{8}$ & $M_{GA}^{10}$   & $M^{10}$ \\
    \midrule
$F1_{train}$ & \textbf{0.879}& 0.858& 0.853 & 0.819& 0.821 & 0.768&0.804&0.765\\
$F1_{test}$ & 0.727& 0.722 & \textbf{0.756} & 0.731& 0.740 & 0.710&0.741&0.712\\
$Fit_{train}$ & \textbf{0.795}& 0.781& 0.717 & 0.693& 0.634 & 0.598&0.563&0.536\\
$Fit_{test}$ & \textbf{0.689}& 0.685& 0.649 & 0.631& 0.581 & 0.557&0.519&0.498\\
    \bottomrule
  \end{tabular}
\end{table}

\begin{table}[h!]
  \caption{Model comparisons ($N_{train} = 20000$, $N_{test} = 7000$, 25 epochs, 15 generations)}
  \label{sample-table}
  \centering
  \begin{tabular}{lllllllll}
    \toprule
Model: & $CM_{GA}^{4}$ & $CM^{4}$ & $CM_{GA}^{6}$  & $CM^{6}$   & $CM_{GA}^{8}$   & $CM^{8}$ & $M_{GA}^{10}$   & $M^{10}$ \\
    \midrule
$F1_{train}$ & \textbf{0.893}& 0.879& 0.866 & 0.796& 0.855 & 0.779&0.825&0.712\\
$F1_{test}$ & 0.727& 0.718& \textbf{0.765} & 0.720 &  0.761 & 0.721&0.744&0.665\\
$Fit_{train}$ & \textbf{0.805}& 0.795& 0.726 & 0.677& 0.659 & 0.605&0.578&0.498\\
$Fit_{test}$ & \textbf{0.689}& 0.683& 0.656 & 0.624& 0.593 & 0.565&0.521&0.466\\
    \bottomrule
  \end{tabular}
\end{table}

\begin{table}[h!]
  \caption{Model comparisons ($N_{train} = 20000$, $N_{test} = 7000$, 20 epochs, 20 generations)}
  \label{sample-table}
  \centering
  \begin{tabular}{lllllllll}
    \toprule
Model: & $CM_{GA}^{4}$ & $CM^{4}$ & $CM_{GA}^{6}$  & $CM^{6}$   & $CM_{GA}^{8}$   & $CM^{8}$ & $M_{GA}^{10}$   & $M^{10}$ \\
    \midrule
$F1_{train}$ & \textbf{0.883}& 0.802& 0.849 & 0.767& 0.835 & 0.803&0.813&0.733\\
$F1_{test}$ & 0.730& 0.691& \textbf{0.750} & 0.709& 0.749 & 0.732&0.744&0.677\\
$Fit_{train}$ & \textbf{0.798}& 0.741& 0.714 & 0.657& 0.645 & 0.622&0.569&0.513\\
$Fit_{test}$ & \textbf{0.691}& 0.664& 0.645 & 0.616& 0.584 & 0.572&0.521&0.474\\
    \bottomrule
  \end{tabular}
\end{table}

\begin{table}[h!]
  \caption{Model comparisons ($N_{train} = 20000$, $N_{test} = 7000$, 25 epochs, 20 generations)}
  \label{sample-table}
  \centering
  \begin{tabular}{lllllllll}
    \toprule
Model: & $CM_{GA}^{4}$ & $CM^{4}$ & $CM_{GA}^{6}$  & $CM^{6}$   & $CM_{GA}^{8}$   & $CM^{8}$ & $M_{GA}^{10}$   & $M^{10}$ \\
    \midrule
$F1_{train}$ & \textbf{0.893}& 0.862& 0.867 & 0.832& 0.852 & 0.785&0.815&0.749\\
$F1_{test}$ & 0.736& 0.714& 0.759 & 0.736& \textbf{0.760} & 0.708&0.741&0.695\\
$Fit_{train}$ & \textbf{0.805}& 0.783& 0.727 & 0.702& 0.656 & 0.610&0.571&0.524\\
$Fit_{test}$ & \textbf{0.695}& 0.680& 0.651 & 0.635& 0.582 & 0.556&0.519&0.487\\
    \bottomrule
  \end{tabular}
\end{table}

\begin{table}[h!]
  \caption{Model comparisons ($N_{train} = 15000$, $N_{test} = 6000$, 20 epochs, 25 generations)}
  \label{sample-table}
  \centering
  \begin{tabular}{lllllllll}
    \toprule
Model: & $CM_{GA}^{4}$ & $CM^{4}$ & $CM_{GA}^{6}$  & $CM^{6}$   & $CM_{GA}^{8}$   & $CM^{8}$ & $M_{GA}^{10}$   & $M^{10}$ \\
    \midrule
$F1_{train}$ & \textbf{0.899}& 0.886& 0.870 & 0.825& 0.837 & 0.760&0.780&0.748\\
$F1_{test}$ & 0.711& 0.714& \textbf{0.741} & 0.725& 0.733 & 0.695&0.693&0.686\\
$Fit_{train}$ & \textbf{0.809}& 0.800& 0.729 & 0.698& 0.646 & 0.592&0.546&0.524\\
$Fit_{test}$ & 0.678& \textbf{0.680}& 0.639 & 0.628& 0.573 & 0.547&0.485&0.480\\
    \bottomrule
  \end{tabular}
\end{table}

\begin{table}[h!]
  \caption{Model comparisons ($N_{train} = 20000$, $N_{test} = 7000$, 20 epochs, 25 generations)}
  \label{sample-table}
  \centering
  \begin{tabular}{lllllllll}
    \toprule
Model: & $CM_{GA}^{4}$ & $CM^{4}$ & $CM_{GA}^{6}$  & $CM^{6}$   & $CM_{GA}^{8}$   & $CM^{8}$ & $M_{GA}^{10}$   & $M^{10}$ \\
    \midrule
$F1_{train}$ & \textbf{0.883}& 0.880& 0.854 & 0.836& 0.835 & 0.772&0.803&0.779\\
$F1_{test}$ & 0.727& 0.732& \textbf{0.762} & 0.743& 0.745 & 0.715&0.740&0.714\\
$Fit_{train}$ & \textbf{0.798}& 0.796& 0.718 & 0.705& 0.645 & 0.600&0.562&0.545\\
$Fit_{test}$ & 0.689& \textbf{0.692}& 0.653 & 0.640& 0.582 & 0.561&0.518&0.500\\
    \bottomrule
  \end{tabular}
\end{table} 

\subsubsection{Simulation results using the second data partition method}

Simulation results using the second data partition method are shown in Tables 14-18.

\begin{table}[h!]
  \caption{Model comparisons ($K_{train} = 20000$, $K_{test} = 7000$, 20 epochs, 5 generations)}
  \label{sample-table}
  \centering
  \begin{tabular}{lllllllll}
    \toprule
Model: & $CM_{GA}^{4}$ & $CM^{4}$ & $CM_{GA}^{6}$  & $CM^{6}$   & $CM_{GA}^{8}$   & $CM^{8}$ & $M_{GA}^{10}$   & $M^{10}$ \\
    \midrule
$F1_{train}$ & \textbf{0.875}& 0.849& 0.840 & 0.819& 0.807 & 0.807&0.772&0.759\\
$F1_{test}$ & 0.725& 0.703& \textbf{0.747} & 0.736& 0.736 & 0.736&0.721&0.708\\
$Fit_{train}$ & \textbf{0.793}& 0.774& 0.708 & 0.693& 0.625 & 0.625&0.540&0.531\\
$Fit_{test}$ & \textbf{0.688}& 0.672& 0.643 & 0.635& 0.575 & 0.575&0.505&0.496\\
    \bottomrule
  \end{tabular}
\end{table}

\begin{table}[h!]
  \caption{ Model comparisons ($K_{train} = 20000$, $K_{test} = 7000$, 20 epochs, 10 generations)}
  \label{sample-table}
  \centering
  \begin{tabular}{lllllllll}
    \toprule
Model: & $CM_{GA}^{4}$ & $CM^{4}$ & $CM_{GA}^{6}$  & $CM^{6}$   & $CM_{GA}^{8}$   & $CM^{8}$ & $M_{GA}^{10}$   & $M^{10}$ \\
    \midrule
$F1_{train}$ & \textbf{0.863}& 0.855& 0.851 & 0.824& 0.817 & 0.801&0.776&0.730\\
$F1_{test}$ & 0.716& 0.715& \textbf{0.765} & 0.734& 0.742 & 0.732&0.720&0.687\\
$Fit_{train}$ & \textbf{0.784}& 0.779& 0.716 & 0.697& 0.632 & 0.621&0.543&0.511\\
$Fit_{test}$ & \textbf{0.681}& \textbf{0.681}& 0.656 & 0.634& 0.579 & 0.572&0.504&0.481\\
    \bottomrule
  \end{tabular}
\end{table}

\begin{table}[h!]
  \caption{ Model comparisons ($K_{train} = 20000$, $K_{test} = 7000$, 20 epochs, 15 generations)}
  \label{sample-table}
  \centering
  \begin{tabular}{lllllllll}
    \toprule
Model: & $CM_{GA}^{4}$ & $CM^{4}$ & $CM_{GA}^{6}$  & $CM^{6}$   & $CM_{GA}^{8}$   & $CM^{8}$ & $M_{GA}^{10}$   & $M^{10}$ \\
    \midrule
$F1_{train}$ & \textbf{0.878}& 0.867& 0.852 & 0.850& 0.830 & 0.748&0.795&0.763\\
$F1_{test}$ & 0.732& 0.723& \textbf{0.761} & 0.756& 0.756 & 0.698&0.734&0.711\\
$Fit_{train}$ & \textbf{0.795}& 0.787& 0.716 & 0.715& 0.641 & 0.583&0.557&0.534\\
$Fit_{test}$ & \textbf{0.692}& 0.686& 0.653 & 0.649& 0.589 & 0.549&0.513&0.498\\
    \bottomrule
  \end{tabular}
\end{table}

\begin{table}[h!]
  \caption{ Model comparisons ($K_{train} = 20000$, $K_{test} = 7000$, 20 epochs, 20 generations)}
  \label{sample-table}
  \centering
  \begin{tabular}{lllllllll}
    \toprule
Model: & $CM_{GA}^{4}$ & $CM^{4}$ & $CM_{GA}^{6}$  & $CM^{6}$   & $CM_{GA}^{8}$   & $CM^{8}$ & $M_{GA}^{10}$   & $M^{10}$ \\
    \midrule
$F1_{train}$ & \textbf{0.878}& 0.850& 0.852 & 0.835& 0.832 & 0.811&0.807&0.785\\
$F1_{test}$ & 0.731& 0.717& \textbf{0.764} & 0.748& 0.755 & 0.739&0.743&0.731\\
$Fit_{train}$ & \textbf{0.795}& 0.775& 0.716 & 0.705& 0.642 & 0.628&0.565&0.550\\
$Fit_{test}$ & \textbf{0.692}& 0.682& 0.655 & 0.644& 0.589 & 0.577&0.520&0.512\\
    \bottomrule
  \end{tabular}
\end{table}

\begin{table}[h!]
  \caption{ Model comparisons ($K_{train} = 20000$, $K_{test} = 7000$, 20 epochs, 25 generations)}
  \label{sample-table}
  \centering
  \begin{tabular}{lllllllll}
    \toprule
Model: & $CM_{GA}^{4}$ & $CM^{4}$ & $CM_{GA}^{6}$  & $CM^{6}$   & $CM_{GA}^{8}$   & $CM^{8}$ & $M_{GA}^{10}$   & $M^{10}$ \\
    \midrule
$F1_{train}$ & \textbf{0.887}& 0.868& 0.857 & 0.804& 0.830 & 0.803&0.804&0.769\\
$F1_{test}$ & 0.730& 0.720& \textbf{0.768} & 0.734& 0.757 & 0.727&0.743&0.715\\
$Fit_{train}$ & \textbf{0.801}& 0.788& 0.720 & 0.683& 0.641 & 0.622&0.563&0.538\\
$Fit_{test}$ & \textbf{0.691}& 0.684& 0.658 & 0.634& 0.590 & 0.569&0.520&0.501\\
    \bottomrule
  \end{tabular}
\end{table}

\subsection{Performance analysis}

\subsubsection{Comparison between GA-based MCNN models and non-GA-based MCNN models}

Simulation results as shown in Tables 4-13 for the first data partition and Tables 14-18 for the second data partition indicated that 4 MCNN models (i.e., $CM_{GA}^{4}$, $CM_{GA}^{6}$, $CM_{GA}^{8}$ and $M_{GA}^{10}$) generated by Algorithm 2 with the new GA  outperformed 4 MCNN models (i.e., $CM^{4}$, $CM^{6}$, $CM^{8}$ and $M^{10}$) without the new GA in terms of testing F1-scores for 58 cases among 60 cases ($CM_{GA}^{8}$ tied with $CM^{8}$ as shown in Tables 7 and 14), respectively. Thus, Algorithm 2 with the new GA is useful. 

\subsubsection{Comparison between CMCNN models and non-compressed MCNN models}
Simulation results as shown in Tables 4-18 indicated that three CMCNN Models with the new GA had lower testing F1-scores and lower memory usage than a non-compressed MCNN Model ($M_{GA}^{10}$) with the new GA for 43 cases among 45 cases ($CM_{GA}^{4}$ had higher testing F1-scores than $M_{GA}^{10}$ as shown in Tables 17 and 18). In addition, three CMCNN Models without the new GA had lower testing F1-scores and lower memory usage than a non-compressed MCNN Model ($M^{10}$) without the new GA for 42 cases among 45 cases ($CM^{4}$ had higher testing F1-scores than $M^{10}$ as shown in Tables 14 and 17, and $CM^{8}$ had higher testing F1-scores than $M^{10}$ as shown in Table 16). Thus, CMCNN models can outperform a non-compressed MCNN Model. 

\subsubsection{Overall comparisons}

The largest value for each row is bolded in Tables 4-18. The best CNN models come from the two smallest CMCNNs in terms of the number of convolutional layers. Also, all bolded testing F1-scores are for CMCNN models using the new GA, showing that the new GA is useful. Although the best CMCNN selected by Algorithm 2 does not have the best testing F1-scores, it has the shortest execution time (smallest power usage) and the smallest model size (454KB, smallest memory usage).

\section{Conclusions}

Simulation results show that CMCNNs can achieve better performance, shorter training time (i.e., less power consumption), and less memory usage (model size) than MCNNs. For example, $CM_{GA}^{4}$ with 454KB model size is more accurate, memory-efficient, power-efficient, and faster than $M^{10}$ with 815KB model size. Results show that the new GA based model selection algorithm can perform better than a random model selection algorithm. $CM_{GA}^{k}$ outperformed $CM^{k}$ for $k=4, 6, 8$ in most simulation cases, so the new GA is an effective method in automatically finding the best CMCNNs. However, the new GA is  prone to overfitting. Effective, fast, power-efficient, and memory-efficient CMCNNs using a small number of convolutional layers with different activation functions can be used in various applications, especially in computer vision. With faster training, computer systems and mobile devices running CMCNNs can save power, which would increase power efficiency and battery life. CMCNNs are more memory-efficient than MCNNs since the model sizes of CMCNNs are much smaller than those of MCNNs. 

\section{Future works}

The fitness function for GA does not have to be linear, and it can be changed and optimized to meet different users’ requirements, especially the weights. The simulation results suggest that the model size may overpower the testing F1-score since the model with the second smallest model size always had the highest testing F1-score. Thus, the weights of the fitness function will be adjusted by a user. Traditional techniques, such as dropout, will be used to control overfitting. Stratied $k$-fold cross validation will be used for better model evaluation. Other compressed deep neural networks, such as ResNets and DenseNets with a reduced number of convolutional layers using different activation functions, will be developed to increase classification accuracy, reduce training times, reduce computational power, and reduce memory usage (model size) for various applications especially for systems with limited power and memory. The optimized power-efficient and memory-efficient CMCNNs may also be pre-trained and embedded into mobile devices. Better algorithms will be created to find the best architecture with the best set of activation functions to build an effective, power-efficient, and memory-efficient CMCNN. Other NAS methods that are not evolution-based search algorithms will be tested. Other general optimization techniques, such as particle swarm optimization [15-16] and microcanonical annealing [17], will be used to develop more effective CMCNN model selection algorithms. Since the CMCNN model selection among a large number of potential CMCNNs takes a very long time for many generations, parallel optimization techniques will be developed. It is highly beneficial to create and implement intelligent and efficient optimization algorithms and software for automatically identifying the best CMCNN for important practical applications, such as biomedical imaging, mobile computing, and brain informatics.

\section*{References}

\medskip

\small

[1] LeCun, Y., Bengio, Y.\ \& Hinton, G.E. (2015) Deep learning. Nature 521, pp.\ 436--444.

[2] Krizhevsky, A., Sutskever, I.\ \& Hinton, G.E.\ (2012) Imagenet classification with deep convolutional neural networks. In {\it Advances in Neural Information Processing Systems 25}, pp.\ 1097--1105. Cambridge, MA: MIT Press.

[3] He, K., Zhang, X., Ren, S.\ \& Sun, J. \ (2016) Deep Residual Learning for Image Recognition. In Proceedings of the 2016 IEEE Conference on Computer Vision and Pattern Recognition (CVPR), pp.\ 770--778.

[4] Esteva, A., Kuprel, B., Novoa, R.A., Ko, J., Swetter, S.M., Blau, H.M.\ \& Thrun, S.\ (2017) Dermatologist-level classification of skin cancer with deep neural networks. {\it Nature} {\bf 542}(7639):115--118.

[5] Szegedy, C., Liu, W., Jia, Y., Sermanet, P., Reed S., Anguelov, D., Erhan, D., Vanhoucke, V.\ \& Rabinovich, A.\ (2015) Going Deeper with Convolutions. In Proceedings of 2015 IEEE Conference on Computer Vision and Pattern Recognition (CVPR), pp.\ 1--9. 

[6] Szegedy, C., Ioffe, S., Vanhoucke, V.\ \& Alemi, A.\ (2017) Inception-v4, Inception-ResNet and the Impact of Residual Connections on Learning. In Proceedings of the Thirty-First AAAI Conference on Artificial Intelligence (AAAI-17), pp.\ 4278--4284.

[7] Sun, Y., Xue, B., Zhang, M. \ \& Yen, G.\ (2018) Automatically Designing CNN Architectures Using Genetic Algorithm for Image Classification. [Online.] Available: https://arxiv.org/pdf/1808.03818.pdf.

[8] You, Z. \ \& Pu, Y. \ (2015) The Genetic Convolutional Neural Network Model Based on Random Sample. International Journal of u- and e- Service, Science and Technology. pp.\ 317--326.

[9] Lu, Z., Whalen, I., Boddeti, V., Dhebar, Y. D., Deb, K., Goodman, E. D. \ \& Banzhaf, W.\ (2018) {{NSGA-NET:} {A} Multi-Objective Genetic Algorithm for Neural Architecture Search. http://arxiv.org/abs/1810.03522. 

[10] Krizhevsky, A.\ (2009) Learning multiple layers of features from tiny images. https://www.cs.toronto.edu/~kriz/learning-features-2009-TR.pdf. 

[11] Molchanov, P., Tyree, S., Karras, T., Aila, T., \& Kautz, J.\ (2016) Pruning convolutional neural networks for resource efficient inference. [Online.] Available: https://arxiv.org/abs/1611.06440.

[12] Li, K.\ (2008) Performance analysis of power-aware task
scheduling algorithms on multiprocessor computers with
dynamic voltage and speed. IEEE Transactions on Parallel
and Distributed Systems, 19(11):1484–1497. 

[13] OASIS Brains Datasets. [Online]. Available:
http://www.oasis-brains.org/\#data.

[14] Marcus, D.S., Wang, T.H., Parker, J., Csernansky, J.G., Morris, J.C. \ \&  Buckner, R.L.\ (2007) Open access series of imaging studies (oasis):
cross-sectional MRI data in young, middle aged, nondemented, and
demented older adults. $Journal$ $of$ $Cognitive$ $Neuroscience$, 19(9):1498--1507.

[15] Sinha, T. , Haidar, A.  \ \&  Verma, B.\ (2018)  
Particle Swarm Optimization Based Approach for Finding Optimal Values of Convolutional Neural Network Parameters. 2018 IEEE Congress on Evolutionary Computation (CEC), pp.\ 1—6.

[16] Tan, T. Y., Zhang, L., Lim, C. P., Fielding, B., Yu, Y. \ \& Anderson,  E.\ (2019)  
Evolving Ensemble Models for Image Segmentation Using Enhanced Particle Swarm Optimization. IEEE Access, vol. 7, pp.\ 34004—34019.

[17] Ayumi, V., Rasdi Rere, L. M., Fanany, M. I. \ \&  Arymurthy, A. M.\ (2016)  Optimization of convolutional neural network using microcanonical annealing algorithm. 2016 International Conference on Advanced Computer Science and Information Systems (ICACSIS), pp.\ 506—3511.

\end{document}